\documentclass[runningheads]{llncs}

\usepackage{eccv}

\usepackage{eccvabbrv}

\usepackage{graphicx}
\usepackage{booktabs}
\usepackage{booktabs,multirow,makecell,array}
\usepackage{xspace}
\usepackage{pifont}
\usepackage{wrapfig}
\usepackage{amsmath,amssymb}
\usepackage{subcaption}
\usepackage[ruled,vlined]{algorithm2e}
\SetAlgoSkip{smallskip}
\SetKwInput{KwIn}{Input}
\SetKwInput{KwOut}{Output}

\usepackage[accsupp]{axessibility}  
\usepackage{hyperref}
\usepackage{orcidlink}

\newcommand{\HSGt}{\textsc{HSG}\textsubscript{$t$}\xspace}
\newcommand{\HSGtO}{$\text{HSG}^{\text{O}}_t \ $}
\newcommand{\HSGtZ}{$\text{HSG}^{\text{Z}}_t \ $}
\newcommand{\HSGtR}{$\text{HSG}^{\text{R}}_t \ $}
\newcommand{\cmark}{\ding{51}} 
\newcommand{\xmark}{\ding{55}} 

\begin{document}

\title{
Hierarchical 3D Scene Graph Construction and Belief-based Planning for Semantic Navigation} 

\titlerunning{Hierarchical Scene Graphs for Semantic Navigation}

\author{
Bing Wu \orcidlink{0000-0001-6175-3944} 
\and
Zuyao Chen \orcidlink{0000-0002-7344-1101} 
\and
Changwen Chen  \orcidlink{0000-0002-6720-234X} 
\thanks{Corresponding author.}
}

\authorrunning{B.~Wu et al.}

\institute{
The Hong Kong Polytechnic University \\
\email{\{bing-comp.wu,zuyao.chen\}@connect.polyu.hk} \\
\email{changwen.chen@polyu.edu.hk}
}

\maketitle

\begin{abstract}
Semantic navigation is a fundamental task for embodied agents operating in unseen environments, requiring both semantic understanding and long-term decision-making. Recent foundation models have empowered agents with rich semantic  priors for this task. However, without structured global representations,  decision-making often falls back on local observations and greedy strategies, resulting in inefficient exploration and myopic behaviors, especially in long-distance navigation. To address these challenges, we propose a zero-shot semantic navigation framework. Our method incrementally maintains an online Hierarchical 3D Scene Graph (HSG) to form a multi-granular semantic topology over objects, zones, and regions, serving as a compact state abstraction for global planning. Building on this memory, we introduce a hierarchical belief-based planning framework that fuses semantic priors with exploration evidence on the HSG, and performs finite-horizon rollouts on an HSG-based simulator to explicitly estimate the long-term expected returns of candidate macro-actions. This enables globally consistent decisions and reduces redundant backtracking. Extensive experiments in high-fidelity simulation environments across multiple tasks and datasets demonstrate that our method outperforms existing state-of-the-art methods, particularly in long-distance scenarios, where our approach improves SR and SPL by an average of 9.4\% and 5.0\%, respectively.

\end{abstract}

\section{Introduction}
\label{sec:intro}
Semantic navigation is not merely about \textit{where an agent can go} but \textit{where it should go}. In unseen environments, an agent must navigate to a semantic target using only egocentric observations under a limited action budget. The target can be specified as an object category (ObjectNav  \cite{ON-nips2020}) or a textual description of a specific instance (TextInstanceNav \cite{InstanceNav}), which requires joint semantic-spatial perception and long-horizon planning. Recent foundation models have advanced open-vocabulary perception and commonsense reasoning, making zero-shot semantic navigation feasible without task-specific retraining \cite{esc_2023_icml}.

\begin{figure}[tb]
  \centering
  \begin{subfigure}{0.5\linewidth}
    \includegraphics[width=0.95\linewidth]{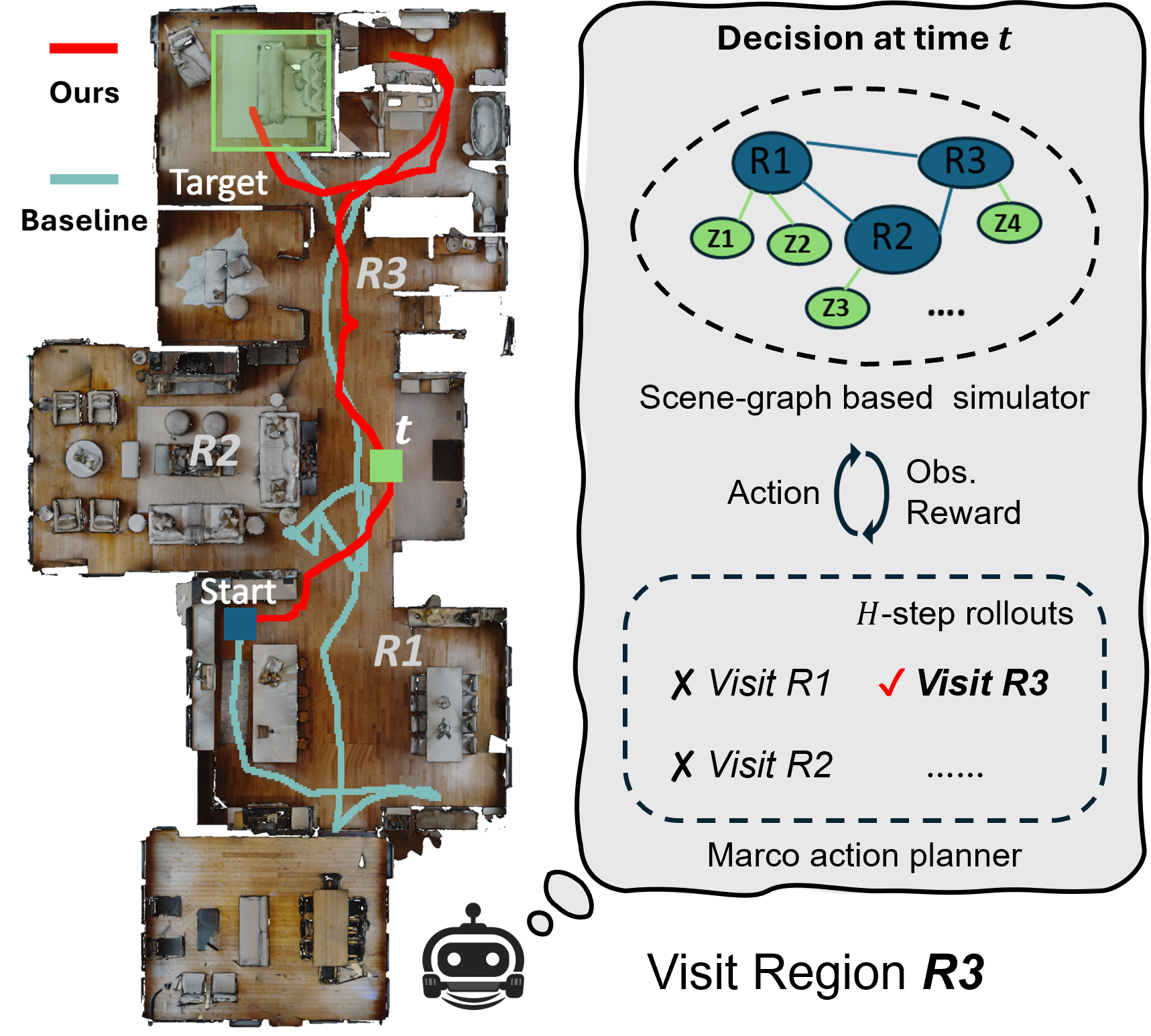}
    \caption{Planning on the HSG.}
    \label{fig:short-a}
  \end{subfigure}
  \hfill
  \begin{subfigure}{0.45\linewidth}
    \includegraphics[width=1.\linewidth]{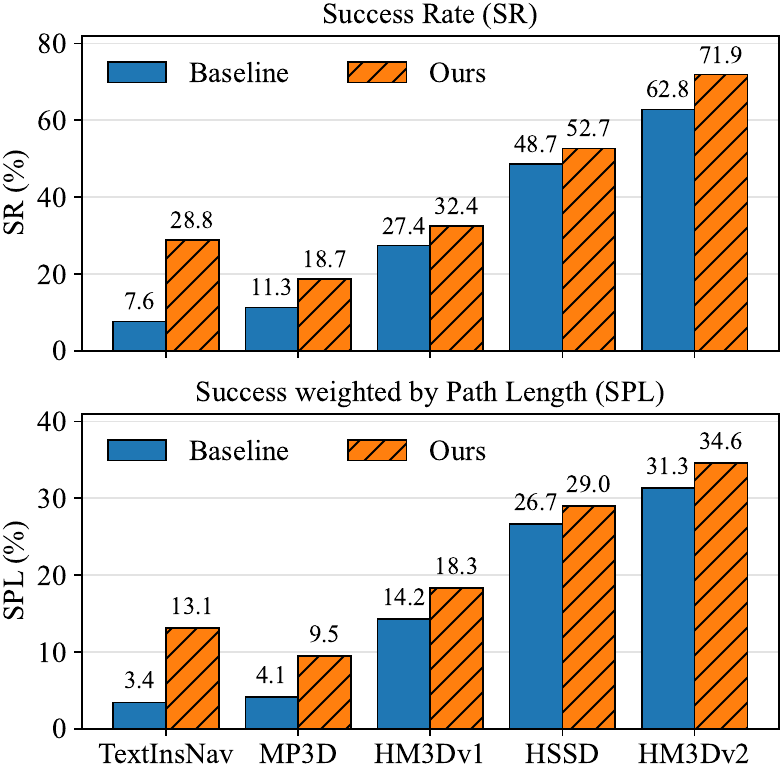}
    \caption{Performance on long-distance navigation.}
    \label{fig:short-b}
  \end{subfigure}
    \caption{
Hierarchical belief-based planning for long-distance navigation. (a) Unlike the greedy exploration of baselines (cyan path), our agent evaluates long-term expected returns via rollouts on the HSG simulator, enabling globally efficient navigation (red path). (b) Our method outperforms state-of-the-art baselines (ApexNav~\cite{apexnav} for ObjectNav, UniGoal~\cite{unigoal} for TextInstanceNav) across multiple datasets in long-distance scenarios.
    }
  \label{fig:short}
\end{figure}

However, despite their powerful perception and commonsense reasoning, foundation models do not directly address the core challenges of spatial memory and long-horizon decision making in semantic navigation \cite{travelplanner}. Most existing zero-shot methods  lack a global structured semantic representation of the environment and instead rely on local observations with greedy or heuristic search strategies \cite{esc_2023_icml,vlfm,sgnav}, which often results in myopic exploration with redundant revisits and frequent backtracking. Moreover, when foundation model priors misalign with the actual scene,  the agent often lacks the explicit observation feedback required to update its belief state. Such errors can accumulate over time, especially in long-distance navigation.

To address these limitations, the agent requires long-horizon planning capabilities, which inherently demand a structured state space with multi-granular semantic abstractions. Recently, Hierarchical 3D Scene Graphs (HSGs) have emerged as a compact and semantically rich environmental representation. However, most existing methods focus on offline or passive reconstruction from complete point clouds \cite{HOVSG} or RGB-D sequences \cite{hughes2024foundations}, assuming full global observability. This offline construction paradigm cannot be directly applied to online exploration tasks constrained by partial observations. Therefore, how to incrementally construct such a hierarchical structure online in unseen environments and integrate it into the decision-making process remains an open challenge.

To bridge this gap, we propose an online hierarchical scene-graph navigation framework that couples \emph{incremental  hierarchical 3D scene graph construction} with \emph{belief-based hierarchical planning}. 
As shown in Fig.~\ref{fig:short-a}, HSG rollouts enable long-horizon planning beyond greedy exploration.
In the perception module, we maintain observed 3D object instances from RGB-D observations and establish spatial adjacency among objects via Generalized Voronoi Diagram (GVD).  Leveraging spectral clustering on combined geometric and semantic features, we incrementally construct a bottom-up, multi-granular semantic topology with an Object–Zone–Region hierarchy (Sec.~\ref{sec:hsg}). In the decision module, we maintain a hierarchical belief state over the \HSGt, which combines large language model (LLM) semantic priors and frontier information gain evidence. We then construct a scene-graph-based simulator to support finite-horizon  lookahead with Partially Observable Monte Carlo Tree Search (POUCT) \cite{pouct}. By simulating state transitions and observation feedback on the HSG, the planner explicitly evaluates the  long-term expected returns of candidate macro-actions to select the optimal action. Finally, a local planner executes the selected macro-action by optimizing the local path and generating low-level commands (Sec.~\ref{sec:policy}).

In short, the contributions of this work can be summarized as follows:
\begin{itemize}
  \item \textit{Online incremental hierarchical 3D scene graph construction.} We propose a bottom-up approach combining GVD and spectral clustering to abstract local observations into a multi-granular semantic topology, providing a structured global memory.
  \item \textit{Hierarchical belief-based planning.} We introduce a planning method that fuses LLM priors with observation feedback. By evaluating long-term expected returns via rollouts on the HSG, it effectively overcomes the myopia of greedy strategies.
  \item \textit{Superior long-distance performance.} Extensive experiments across multiple tasks and datasets demonstrate that our method significantly outperforms state-of-the-art approaches. Most notably, in long-distance episodes, we improve SR and SPL by an average of $9.4\%$ and $5.0\%$, respectively.
\end{itemize}

\section{Related Work}
\textbf{3D Scene Graph Representation.}
Compared to traditional geometric maps (\eg, point clouds, voxels) or topological maps, 3D scene graphs have recently emerged as a compact representation that jointly encodes geometry, semantics, and relational structure in 3D environments \cite{3dsg-iccv2019}. Building on this idea, robotic systems such as Kimera \cite{rosinol2021kimera} and Hydra \cite{hughes2024foundations} couple SLAM with semantic parsing to construct 3D scene graphs online, allowing an agent to maintain a sparse, object-centric map of the environment rather than only a dense metric map. More recent works \cite{open3dsg,conceptgraphs}  further explore hierarchical and open-vocabulary 3D scene graphs. In hierarchical formulations, nodes are organized into multiple abstraction levels, \eg, building, floor, room, and object, so that the environment is described at different granularities within a unified structure, which in turn supports richer and more flexible scene understanding. Once such a 3D scene graph has been constructed, it can serve as a powerful substrate for complex semantic tasks and long-horizon planning, as demonstrated by systems such as SayPlan \cite{sayplan} and HOVSG \cite{HOVSG}.

Despite this progress, many existing hierarchical and open-vocabulary approaches focus on offline reconstruction of 3D scene graphs from complete point clouds \cite{HOVSG} or RGB-D sequences \cite{conceptgraphs}. While systems such as Kimera \cite{rosinol2021kimera} and Hydra \cite{hughes2024foundations} support online construction, they primarily focus on SLAM and scene reconstruction rather than using the graph as a planning state for semantic navigation. This motivates our online hierarchical 3D scene graph, which is incrementally constructed from RGB-D observations and explicitly tailored to support semantic-navigation decision making.

\begin{figure*}[t]
    \centering
    \includegraphics[width=1.0\linewidth]{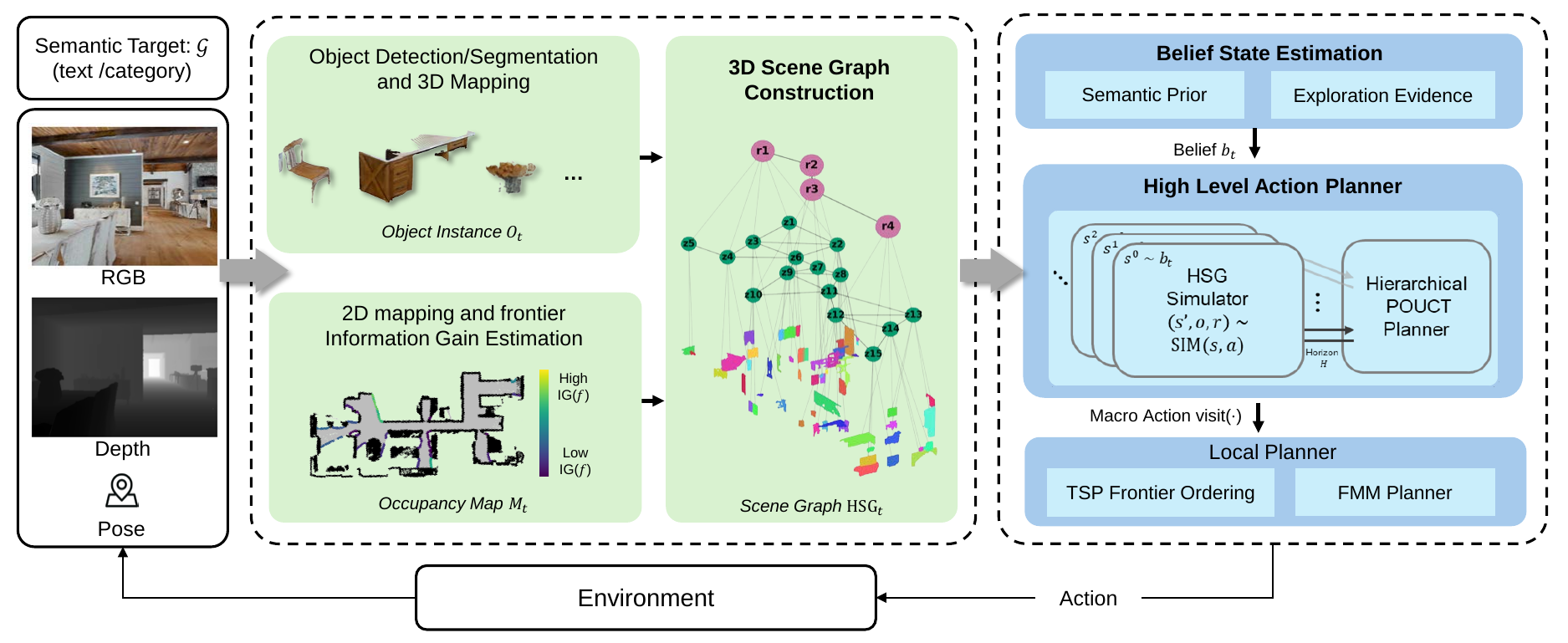}
    \caption{
    Framework of our method. Given posed RGB-D observations, the perception module incrementally builds an HSG along with object instances and occupancy map, which the decision module uses to maintain a belief state $b_t$, select visit macro-actions via hierarchical POUCT, and generate executable actions with local planning.
    }
    \label{fig:pipeline}
\end{figure*}

\noindent \textbf{Semantic Navigation.}
Semantic navigation requires robots to reason over \emph{semantic goals} beyond geometric obstacle avoidance. 
Early Object Navigation (ObjectNav) trains policies at scale in simulation under a closed set of target categories \cite{ON-nips2020}, but such policies depend on task-specific data and reward design and often generalize poorly to novel environments and unseen categories \cite{cows-cvpr2023}.
With open-vocabulary vision–language representations (\eg, CLIP) and large language / vision-language models (LLMs/VLMs), recent methods (\eg, ESC \cite{esc_2023_icml}, OpenFMNav \cite{OpenFMNav}, VLFM \cite{vlfm}) move toward zero-shot and open-vocabulary navigation, allowing agents to pursue arbitrary targets without retraining.
In zero-shot ObjectNav, a common paradigm is to use an open-vocabulary visual encoder to map the target description and observations into a shared embedding space, and couple it with frontier-based exploration \cite{apexnav} or map-based search strategies \cite{voronav} for navigation; furthermore, LLMs are often queried for commonsense priors over likely target locations \cite{instructnav,cognav}. 

To provide structural context for these priors, recent methods like SG-Nav \cite{sgnav} and UniGoal \cite{unigoal} introduce 3D scene graphs in navigation tasks, using them as structured prompts for LLMs to score and select candidate frontiers. In contrast, we formulate the hierarchical scene graph as a compact state abstraction, enabling us to maintain an online belief state and perform explicit long-horizon lookahead via simulator rollouts.

\section{Problem Definition}
We study the semantic navigation task, where a robot navigates in an unknown 3D indoor environment equipped with an egocentric RGB-D camera and odometry. At each time step, the robot observes the current RGB-D frame and its own pose, and selects an action from a discrete set \{\texttt{move\_forward}, \texttt{turn\_left}, \texttt{turn\_right}, \texttt{stop}\}. An episode is considered successful if the robot reaches a target object $\mathcal{G}$ within at most $T$ steps and issues \texttt{stop} when its distance to $\mathcal{G}$ falls within a fixed threshold. A \emph{semantic target} refers to a textual description of the desired object, capturing its category, intrinsic attributes, and relations to nearby objects; when this description reduces to a category label only, the task degenerates to standard ObjectNav \cite{ON-nips2020}.

\section{Incremental Hierarchical 3D Scene Graph Construction}
\label{sec:hsg}
In this section, we detail the incremental construction of the hierarchical 3D scene graph \HSGt from RGB-D observations. Specifically, while object nodes are actively maintained online, higher-level zone and region nodes are induced in a bottom-up manner leveraging GVD and spectral clustering. Furthermore, each node is enriched with a semantic description generated by a vision-language model (VLM).
An overview of the complete pipeline is shown in Fig.~\ref{fig:pipeline}.

\subsection{Hierarchical 3D Scene Graph}
We define the hierarchical 3D scene graph at time $t$ as
\begin{equation}
\HSGt = (V_t, E_t),
\end{equation}
where
\begin{equation}
V_t = O_t \cup Z_t \cup R_t, 
\end{equation}
denotes the set of Object, Zone, and Region nodes at time $t$. Each node $v \in V_t$ carries a set of attributes, including its 3D point cloud, observation timestamps, and semantic description. The edge set is given by 
\begin{equation}
E_t = E^{\text{intra}}_t \cup E^{\text{inter}}_t, 
\end{equation}
which captures two types of relations. Intra-layer spatial relations $E^{\text{intra}}_t$ connect adjacent nodes within the same level, while inter-layer hierarchical relations $E^{\text{inter}}_t$ encode cross-level containment through directed object-to-zone and zone-to-region edges. This hierarchical topology facilitates multi-scale planning across object, zone, and region levels.

\subsection{Geometric Mapping and Object-Level Representation}
At each time step $t$, the agent uses the current RGB-D observation  and pose to incrementally update an occupancy map $M_t$, which represents the explored free space and obstacles. $M_t$ is then used for generalized Voronoi partitioning and low-level action generation. At the same time, following \cite{conceptgraphs}, we apply an open-vocabulary object segmentation model \cite{yoloe} on the RGB image, project the resulting 2D object instances into 3D using depth and pose, and associate them with existing objects to maintain a set of object nodes
\begin{equation}
O_t = \{o_1, \dots, o_{N_t}\},
\end{equation}
where each $o_i$ denotes a 3D object instance in the scene.

\subsection{Hierarchical Layer Construction}

Based on the object instances $O_t$ and the occupancy grid map $M_t$ at time $t$, we first perform object-level Voronoi partitioning to obtain spatial adjacency among objects. We then apply spectral clustering on the object graph, combining geometric proximity and semantic similarity, to induce Zone and Region nodes in a bottom-up manner.

\begin{figure}[t]
\centering
\includegraphics[width=1.\linewidth]{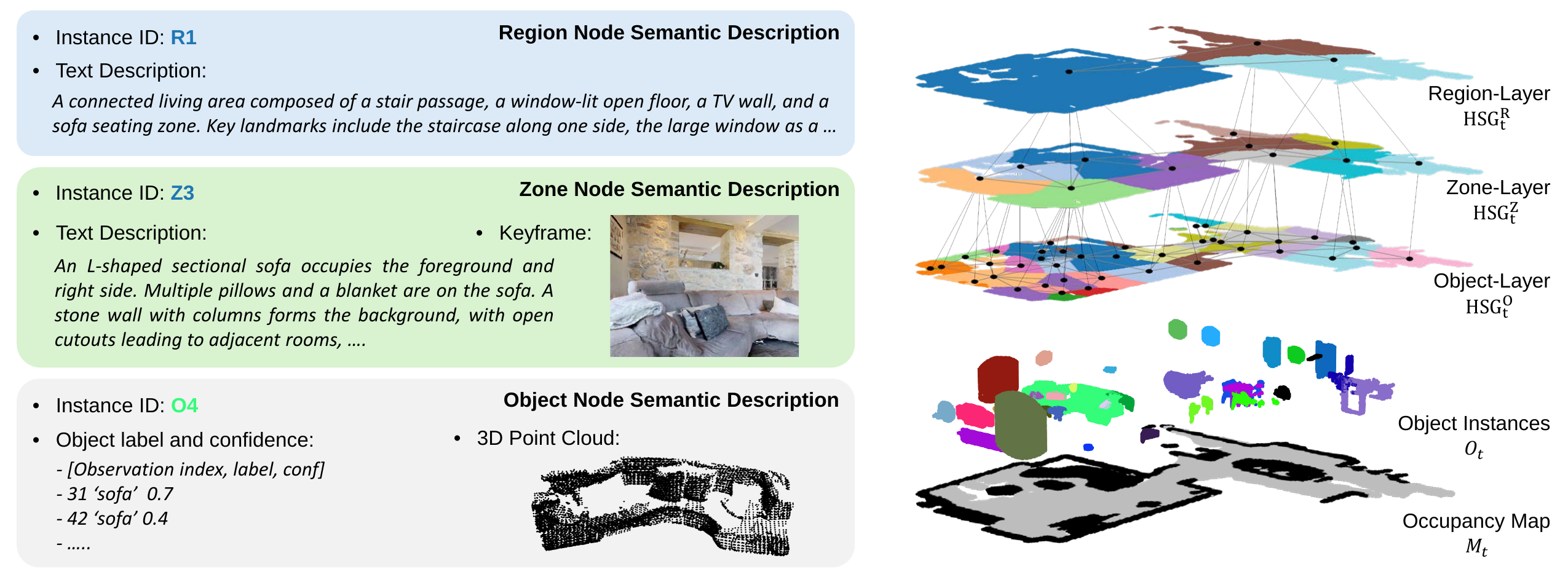}
\caption{
Hierarchical 3D scene graph overview. Given the occupancy map $M_t$ and object instances $O_t$, we construct a bottom-up spatial topology comprising object, zone, and region layers. A GVD partitions the map $M_t$ into object-centric cells, which aggregate into larger-scale zone and region partitions based on their hierarchical dependencies.
}
\label{fig:HSG_vis}
\end{figure}

\noindent \textbf{Object-level Voronoi Partition.}
To establish object adjacency relations, we construct a generalized Voronoi diagram \cite{vor_partition} on the occupancy map. Specifically,  we project the 3D point cloud of each object onto the grid map $M_t$ and obtain a set of occupied cells, which serves as the object’s 2D projection. Each explored grid cell is then assigned to its nearest object based on the shortest path distance to all object projections on the map.  This partitions the explored map into non-overlapping, object-centric areas. Based on this partition, we build an object-level adjacency graph \HSGtO, where each object is represented as a node.  An undirected edge is established between two nodes if their corresponding Voronoi regions share a boundary on the grid.

\noindent  \textbf{Spectral Clustering to Build Zone and Region Nodes.}
We leverage a spatial coherence prior that \emph{nearby objects tend to form coherent higher-level semantic groups}. Based on the object adjacency graph \HSGtO, we cluster objects into higher-level structures by jointly leveraging geometric and semantic cues. Specifically, we construct a weighted affinity matrix $W=[w_{ij}]$ over the adjacency edge set $E^O_t$, where each edge weight encodes spatial proximity and semantic similarity:
\begin{equation}
w_{ij}=
\begin{cases}
\exp\!\left(-\dfrac{d_{ij}^2}{\sigma_d^2}\right)\,
\exp\!\left(\dfrac{\mathrm{sim}(e_i,e_j)}{\tau}\right), & (i,j)\in E^O_t \\
0, & \text{otherwise}
\end{cases},
\end{equation}
where $d_{ij}$ denotes the spatial distance between objects $o_i$ and $o_j$, and $e_i$ represents the semantic embedding encoded from the semantic description of the node. $\mathrm{sim}(\cdot,\cdot)$ is cosine similarity, and $\sigma_d,\tau$ are hyper-parameters. This formulation restricts interactions to spatial neighbors while encouraging semantically coherent grouping. 

To extract the hierarchical structure, the clustering is formulated as a graph cut problem, partitioning $\text{HSG}^{\text{O}}_t$ into subgraphs with \emph{dense internal and sparse inter-cluster connections}. This partitioning is approximately solved via spectral clustering \cite{vor_cluster}. Objects sharing the same cluster label are merged to form a Zone node $z_j \in Z_t$, connected by a hierarchical edge $(o_i, z_j)$. As shown in Fig.~\ref{fig:HSG_vis}, the spatial extent of a Zone is derived from the union of its members' Voronoi areas.

Similarly, we apply the same clustering process to the zone layer \HSGtZ to construct the region layer \HSGtR, which provides a coarser-grained semantic abstraction. The numbers of zones and regions are set according to the currently observed area and the number of observed objects.

\noindent \textbf{Incremental Scene Graph Update.} We adopt a fixed-step incremental update scheme and update the hierarchical scene graph every $N$ steps. For each layer of \HSGt, we maintain a sparse adjacency list representation with edge weights. Within the window $[t-N, t]$, when objects are added or merged, edge weights are recomputed only for the affected objects and their local neighborhoods. Afterward, bottom-up re-clustering is performed on the updated object graph to update the layers.

\subsection{Semantic Description of Scene Graph Nodes}
We maintain a semantic description for each node in \HSGt to derive the semantic prior in Sec.~\ref{sec:policy}. For each object node $o_i$, predicted labels and confidence scores from the open-vocabulary segmentation model are accumulated to obtain stable object semantics, following \cite{apexnav}.  For each zone node, a representative keyframe is selected from past RGB observations by maximizing its coverage of the zone area, and a VLM is used to generate a caption as the zone description. For each region node, descriptions of its member zones are aggregated and summarized by an LLM to produce a coarser region  semantic representation. Furthermore, these semantic descriptions are utilized to determine the \texttt{stop} action. While ObjectNav relies on the object detection results, TextInstanceNav prompts a VLM with the current semantic description and observation images.

\section{Hierarchical Semantic Navigation Policy}
\label{sec:policy}
In this section, we generate the agent’s actions from \HSGt and the occupancy map $M_t$ (Sec.~\ref{sec:hsg}), using a hierarchical policy with high-level decision making and low-level execution. At each high-level step, we maintain a belief state $b_t$ on \HSGt and instantiate a scene-graph-based simulator $\mathrm{SIM}$. Before executing any real actions, $\mathrm{SIM}$ performs forward simulations to estimate the long-horizon return of candidate macro-actions, enabling online selection of the next region or zone to visit. The local planner then converts the selected macro-action into a feasible sequence of frontier visits and discrete actions.

\subsection{Belief State Estimation}
\label{sec:policy_belief_state}

\begin{figure}[t]
\centering
\includegraphics[width=0.9\linewidth]{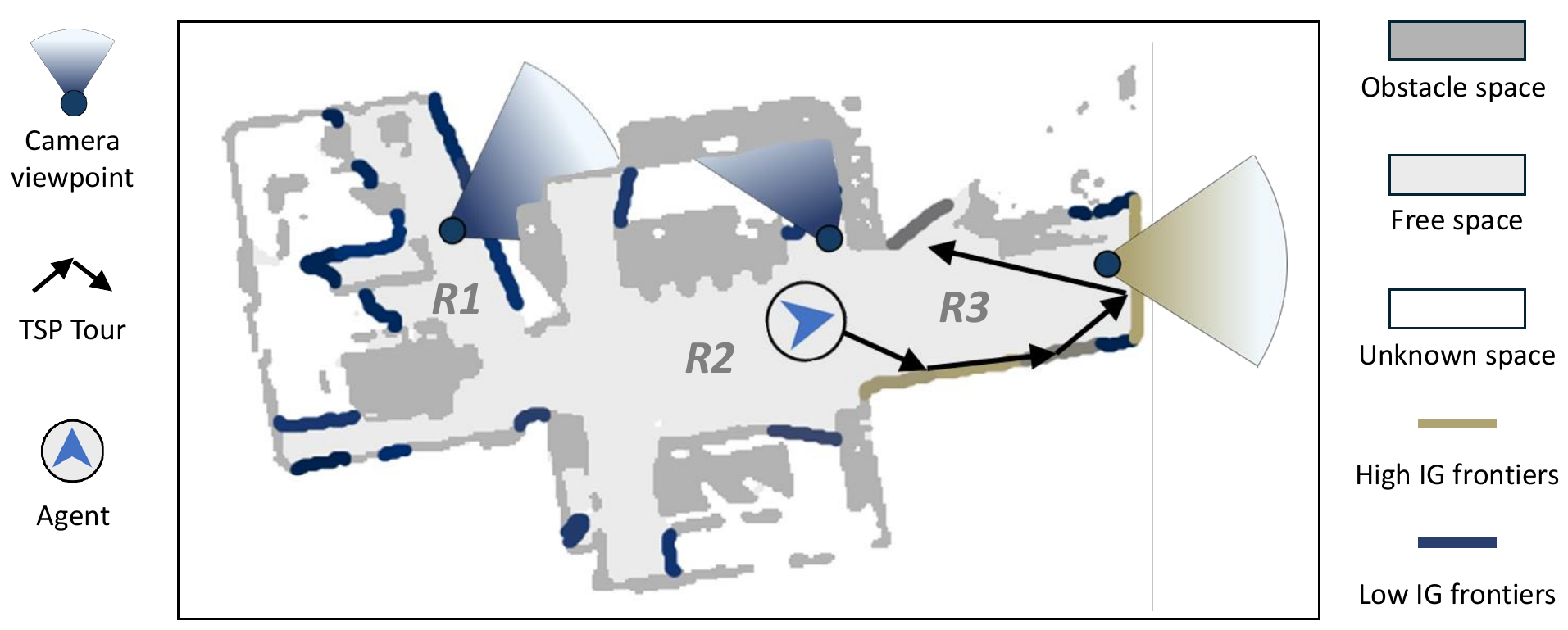}
\caption{
Exploration evidence and local planning. We compute frontier information gains within partitioned regions and aggregate them as exploration evidence to update the HSG belief state. Upon selecting a macro-action, the local planner generates an efficient exploration tour by modeling the visitation order of high-gain frontiers as a traveling salesman problem.
}
\label{fig:info_ig}
\end{figure}

We estimate a hierarchical belief on \HSGt to model the target existence probability  across regions and zones. To avoid overconfidence in explored areas, we introduce virtual nodes $r_u$ and $z_u$ to represent unexplored regions and zones.

\noindent  \textbf{Semantic Prior on the HSG.}
We query an LLM to predict a semantic prior $p_t^{R}(r)$ for each region $r$, reflecting the semantic plausibility that the target appears in that region. For the unknown region  $r_u$, we up-weight its prior when all explored regions receive low scores, and down-weight it otherwise. When the agent enters a region $r$ for refinement, we build local semantic descriptions for each $z\in\mathcal Z_t(r)$ and obtain the zone-level semantic prior $p^{Z}_t(z\mid r)$.

\noindent  \textbf{Observation Feedback as Exploration Evidence.}
We extract frontiers from the occupancy map and compute the information gain $\text{IG}(f)$ for each frontier $f$, which describes its potential to expand the observable free space \cite{frontierIG}, as shown in Fig.~\ref{fig:info_ig}. For region $r$, exploration evidence is defined as

\begin{equation}
e^R_t(r)=
\frac{
\sum\limits_{\;f\in \mathcal F(r)\;:\;\mathrm{IG}(f)>\tau_{\mathcal{G}}} \mathrm{IG}(f)
}{
\sum\limits_{f\in \mathcal F(r)} \mathrm{IG}(f) + \varepsilon
},
\end{equation}
where $\mathcal F(r)$ denotes the set of frontiers in region $r$. This score measures the exploration potential of $r$. If high $IG$ frontiers dominate, $e_t^R(r)\to 1$; otherwise, $e_t^R(r)\to 0$. We set $\tau_{\mathcal{G}}$ by target scale (smaller for small targets, \eg, plant; larger for large targets, \eg, bed), and use $\varepsilon$ for numerical stability.

\noindent \textbf{Belief Construction.}
At most steps, the agent does not find the target and thus receives a $\texttt{not\_found}$ feedback. So we avoid an explicit Bayesian belief update. Instead, we treat this negative signal as a decay of the exploration evidence $e^R_t(r)$ and re-estimate the belief at each high-level step. The region level belief is
\begin{equation}
b_t^R(r) = \text{Norm}\left(  p_t^{R}(r)\cdot e_t^R(r) \right).
\end{equation}

Incorporating observation feedback helps mitigate semantic bias from the LLM. The zone-level belief $b_t^Z(z\mid r^*)$ is constructed analogously.

\begin{algorithm}[t]
\small
\LinesNumbered
\caption{Hierarchical Scene-Graph Belief Planning}
\label{alg:hbp}

\KwIn{Scene graph belief $b_t$ (Sec.~\ref{sec:policy_belief_state}), Simulator $\mathrm{SIM}$ (Sec.~\ref{sec:policy_simulator}),  Horizon $H$}
\KwOut{High-level action $a_t$}

$r^* \leftarrow \texttt{POUCT-Select}(\mathrm{SIM}, b_t^R, \mathcal{A}_t^R, H)$. (Sec.~\ref{sec:policy_pouct})

\If{$r^* = r_u$}{
    \Return $\texttt{visit}(r_u)$  \tcp*[f]{global search}
}

$z^* \leftarrow \texttt{POUCT-Select}(\mathrm{SIM}, b_t^Z(\cdot \mid r^*), \mathcal{A}_t^Z, H)$. (Sec.~\ref{sec:policy_pouct})


\If{$z^* = z_u$}{
    \Return $\texttt{visit}(r^*)$ \tcp*[f]{region-level search}
}
\Else{
    \Return $\texttt{visit}(z^*)$ \tcp*[f]{zone-level search}
}

\end{algorithm}


\subsection{HSG Simulator Construction}
\label{sec:policy_simulator}

We build a scene-graph-based simulator $\text{SIM}$ from the $\mathrm{HSG}_t$ to model state transitions and observation feedback induced by the macro-actions, and use it in Sec.~\ref{sec:policy_pouct} to estimate their long-term returns. Formally, 
\begin{equation}
(s',o, u)\sim \text{SIM}(s,a; \text{HSG}_t). 
\end{equation}

The state $s$ includes the $\text{HSG}_t$, the agent's current node, and a hypothesis of the target node sampled from the current belief $b_t$. A macro-action $a$ corresponds to visiting a region or a zone. 
To keep planning computationally tractable, the topology of $\mathrm{HSG}_t$ is held fixed during each POUCT search.
Executing action $a$ transitions the agent to the selected node and yields a binary observation $o \in \{\texttt{found}, \texttt{not\_found}\}$. If $o = \texttt{not\_found}$, the agent updates its position to explore other nodes in subsequent simulation steps.  The immediate reward $u$ is
\begin{equation}
u(o,s,a)=R(o)-\lambda\,c_{\mathrm{move}}(s,a),\qquad
R(o)=
\begin{cases}
R_{\mathrm{succ}}, & o=\texttt{found} \\
0, & o=\texttt{not\_found}
\end{cases},
\end{equation}
where $c_{\mathrm{move}}(s,a)$ is the shortest-path length on $\mathrm{HSG}_t$ from the agent’s current node to the action’s goal node, and $\lambda$ weights the movement cost.

\subsection{Online Planning with the HSG Simulator}
\label{sec:policy_pouct}

At each high-level decision step $t$, we first select a macro-action $a_t$ over the region belief $b_t^R$ from the region-level action set $\mathcal{A}_t^R$:
\begin{equation}
a_t=\arg\max_{a\in\mathcal A_t^R} Q(b_t^R,a),\qquad
\mathcal A_t^R= \{\texttt{visit}(r)\mid r\in\mathcal R_t \}.
\end{equation}

To estimate $Q(b_t^R,a)$ online, we use POUCT to perform Monte Carlo Tree Search in the current region belief $b_t^R$ (see \cite{pouct} for details on POUCT). Specifically, the $i$-th simulation starts by sampling an initial state instance
\begin{equation}
s^i \sim b_t^R,
\end{equation}
where $s^i$ denotes the state drawn from the belief, specifying the agent’s current node and a target position.
We then perform forward simulations to obtain observations $o$ and immediate rewards $u$ while propagating the state, and back up the accumulated returns to update action-value estimates in the search tree, until reaching the planning horizon $H$ or terminating when $\texttt{found}$ is observed.

Within the search tree, POUCT uses the standard UCB rule to balance exploration and exploitation \cite{pouct}. At leaf nodes, it applies a random policy to extend the trajectory and estimate returns. By repeatedly sampling $s\sim b_t^R$ and backing up accumulated returns over simulated trajectories, POUCT estimates the expected action value as
\begin{equation}
Q(b_t^R,a) \approx \mathbb{E}_{s\sim b_t^R}\Big[\sum_{k=0}^{H-1}\gamma^k u_{t+k}\ \Big|\ a_t=a\Big],
\label{eq.q_value}
\end{equation}
where the planning horizon $H$ controls the look-ahead depth. A larger $H$ accumulates multi-step movement costs and better captures the long-term return of an action. However, an overly large $H$ increases model bias, since $\mathrm{HSG}_t$ is an approximation of the true state and long-horizon simulation accumulates errors, leading to value-estimation drift. Thus, $H$ trades off long-term return evaluation against accumulated model bias.

After obtaining the region-level selection $r^*$, we refine the decision hierarchically. If $r^*=r_u$, we execute a global frontier tour for coverage. Otherwise, we construct a zone-level belief $b_t^Z(\cdot\mid r^*)$ on the subgraph of $r^*$ and run POUCT again over the corresponding zone-level action set
\begin{equation}
\mathcal A_t^Z(r^*)= \{\texttt{visit}(z)\mid z\in\mathcal Z_t(r^*) \},
\end{equation}
 to select the zone to prioritize within the region, as shown in Alg.~\ref{alg:hbp}.

\subsection{Local Planner} 
\label{sec:policy_local_planner}

Given a macro-action $\texttt{visit}(\cdot )$, we collect the set of frontiers associated with the selected region or zone on the occupancy map $M_t$. Following FUEL \cite{fuel}, we cast the ordering of these frontier visits as an asymmetric traveling salesman problem (ATSP) and solve for a minimum-cost tour to reduce local travel distance. Conditioned on the resulting frontier order, we plan shortest paths segment-by-segment on $M_t$ using Fast Marching Method \cite{fmm}, and discretize the concatenated path into a sequence of executable low-level actions.

\section{Experiment}

\begin{table}[t]
\caption{State-of-the-art comparison on the ObjectNav task across four benchmark datasets. ``TF'' indicates training-free methods. Best results are in bold.}
\centering
\resizebox{0.95\textwidth}{!}{
\begin{tabular}{lccccccccc}
\toprule
\multirow{2}{*}{\textbf{Method}} & \multirow{2}{*}{\textbf{TF}} & \multicolumn{2}{c}{\textbf{MP3D}} & \multicolumn{2}{c}{\textbf{HM3Dv1}} & \multicolumn{2}{c}{\textbf{HM3Dv2}} & \multicolumn{2}{c}{\textbf{HSSD}} \\ 
\cmidrule(lr){3-4} \cmidrule(lr){5-6} \cmidrule(lr){7-8} \cmidrule(lr){9-10}
 &  & SR $\uparrow$ & SPL $\uparrow$ & SR $\uparrow$ & SPL $\uparrow$ & SR $\uparrow$ & SPL $\uparrow$ & SR $\uparrow$ & SPL $\uparrow$ \\ 
\midrule
ZSON \cite{zson}              & \xmark  & 15.3 & 4.8   & 25.5 & 12.6  & -    & -     & -    & -     \\
Uni-NaVid \cite{uni-navid}           & \xmark  & -    & -     & 73.7 & 37.1  & -    & -     & -    & -     \\ 
VLingNav \cite{vlingnav}             & \xmark  & 58.9 & 26.5    & 79.1 & 42.9  & 83.0  & 40.5    & -    & -     \\

\midrule
ESC \cite{esc_2023_icml}         & \cmark & 28.7 & 14.2  & 39.2 & 22.3  & -    & -     & 38.1 & 22.2  \\
L3MVN \cite{l3mvn}                & \cmark & 34.9 & 14.5  & 50.4 & 23.1  & 36.3 & 15.7  & 41.2 & 22.5  \\
VoroNav \cite{voronav}           & \cmark & -    & -     & 42.0 & 26.0  & -    & -     & 41.0 & 23.2  \\
VLFM \cite{vlfm}                 & \cmark & 36.4 & 17.5  & 52.5 & 30.4  & 63.6 & 32.5  & -    & -     \\
SG-Nav \cite{sgnav}              & \cmark & 40.2 & 16.0  & 54.0 & 24.9  & 49.6 & 25.5  & -    & -     \\
ImagineNav \cite{imaginenav}      & \cmark & -    & -     & 53.0 & 23.8  & -    & -     & 51.0 & 24.9  \\
UniGoal \cite{unigoal}            & \cmark & 41.0 & 16.4  & 54.5 & 25.1  & -    & -     & -    & -     \\ 
BeliefMapNav \cite{beliefmapnav}  & \cmark & 37.3 & 17.6  & 61.4 & 30.6  & -    & -     & 65.2 & 32.1 \\
ImagineNav++ \cite{imaginenav++}  & \cmark &  -   &  -    & 58.5 & 26.6 & -    & -     & 64.5 & 27.9 \\
FBN-Nav \cite{fbnnav}             & \cmark & 42.1 & 18.1  & 58.8 & 31.2 & -    & -     & - & - \\
ApexNav \cite{apexnav}           & \cmark & 39.2 & 17.8  & 59.6 & 33.0  & 76.2 & 38.0  & 61.1  & 31.5 \\
\midrule
\textbf{Ours}                    & \cmark & \textbf{45.9} & \textbf{19.6}  &\textbf{61.6} & \textbf{33.9} & \textbf{80.1} & \textbf{39.5} & \textbf{69.9} & \textbf{36.4} \\ 
\bottomrule
\end{tabular}
}
\label{tab:comparison}
\end{table}

In this section, we present extensive experiments to evaluate our method. We first introduce the experimental setup, then compare our approach with state-of-the-art baselines and perform ablations on its key components. Finally, we provide qualitative results to further illustrate the benefits of our design.

\subsection{Experiment Setup}

\noindent \textbf{Dataset}: We evaluate our approach on two navigation tasks. For Object Navigation (ObjectNav), we follow SemExp \cite{ON-nips2020} and conduct experiments on Matterport3D (MP3D \cite{mp3d}), HM3D-Semantics-v0.1 (HM3Dv1 \cite{hm3dv1}), HM3D-Semantics-v0.2 (HM3Dv2 \cite{hm3dv2}) and Habitat Synthetic Scenes Dataset (HSSD \cite{hssd}). For Text-Instance Navigation, we use the dataset released in InstanceNav \cite{InstanceNav} and follow its original experimental settings.

\noindent \textbf{Evaluation Metrics}: We adopt \textit{Success Rate} (SR) and \textit{Success weighted by Path Length} (SPL) as the evaluation metrics \cite{nav_evaluation}. SR measures the fraction of episodes in which the agent successfully finds the target. SPL measures path efficiency relative to the shortest path, and is set to 0 for failed episodes.

\noindent \textbf{Implementation Details}:  We implement our agent in Habitat-Sim \cite{habitat3.0}. We use Qwen3-VL-8B \cite{qwen3} as both the vision-language model and the language model to generate semantic descriptions of the scene graph and to estimate semantic priors over target objects. For open-vocabulary object detection and segmentation, we adopt YOLOE \cite{yoloe}. All simulation experiments are run on a workstation with an NVIDIA GeForce RTX 4090 GPU and an Intel Core i9-12900K CPU.
The map and HSG are updated online, although VLM/LLM inference prevents real-time execution; detailed runtime analysis is provided in the supplementary material.

\noindent \textbf{Baselines}:
We evaluate our method on ObjectNav and TextInstanceNav against both training-based (as supervised references) and training-free (our primary focus) baselines. While leading training-free methods like VLFM \cite{vlfm}, UniGoal \cite{unigoal} and ApexNav \cite{apexnav} leverage foundation model priors, they lack global multi-granularity memory and explicit long-horizon planning, typically defaulting to local greedy frontier search. Additionally, we select ApexNav as a strong training-free baseline. It adaptively switches between semantic-prior guidance and geometry-driven coverage exploration, and serves as a competitive frontier-based reference. We further compare performance across different target distances.

\subsection{Comparison with State-of-the-Art}

\begin{table}[t]
\scriptsize
\centering

\begin{minipage}[t]{0.49\textwidth}
\caption{Results on Text-instance navigation.}
\centering
\begin{tabular}{
  >{\raggedright\arraybackslash}m{0.35\columnwidth}
  >{\centering\arraybackslash}m{0.1\columnwidth}
  >{\centering\arraybackslash}m{0.18\columnwidth}
  >{\centering\arraybackslash}m{0.18\columnwidth}
}
\toprule
\textbf{Method} & \textbf{TF} & \textbf{SR $\uparrow$} & \textbf{SPL $\uparrow$} \\
\midrule
PSL \cite{InstanceNav}        & \xmark  & 16.5 & 7.5  \\
GOAT \cite{goat}              & \xmark  & 17.0 & 8.8  \\
UniGoal \cite{unigoal}        & \cmark & 20.2 & 11.4 \\
\midrule
Ours                           & \cmark & \textbf{38.1} & \textbf{16.0} \\
\bottomrule
\end{tabular}
\label{tab:textnav_hm3d}
\end{minipage}
\hfill
\begin{minipage}[t]{0.49\textwidth}
\caption{Ablations on HM3Dv2 ObjectNav.}
\centering
\begin{tabular}{
  >{\raggedright\arraybackslash}m{0.45\columnwidth}
  >{\centering\arraybackslash}m{0.18\columnwidth}
  >{\centering\arraybackslash}m{0.18\columnwidth}
}
\toprule
\textbf{Method} & \textbf{SR $\uparrow$} & \textbf{SPL $\uparrow$} \\
\midrule
w/o Evidence                                 & 76.0 & 37.4 \\
w/o Rollouts ($H$=0)                         & 76.7 & 35.9 \\
w/o Zone-level                               & 79.5 & 38.2 \\
w/ Qwen3-VL-4B                                & 77.2 & 37.9 \\
\midrule
Ours                                  & \textbf{80.1} & \textbf{39.5} \\
\bottomrule
\end{tabular}
\label{tab:objnav_hm3dv2_ablation_components}
\end{minipage}
\end{table}

We evaluate our approach against state-of-the-art baselines on ObjectNav and TextInstanceNav. On ObjectNav (Tab.~\ref{tab:comparison}), our method improves average SR and SPL by $5.4\%$ and $2.3\%$ compared to the strong baseline ApexNav across four datasets. Gains are even more pronounced on TextInstanceNav (Tab.~\ref{tab:textnav_hm3d}), where we outperform the current state-of-the-art method by $17.9\%$ in SR and $4.6\%$ in SPL. We attribute this to the rich target attributes and spatial relations in TextInstanceNav queries, which fully exploit our HSG for multi-granularity semantic modeling and explicit long-horizon planning capabilities.

\subsection{Long-Distance Navigation Analysis}
To validate our contributions to long-horizon decision-making and spatial memory, we specifically evaluate our method on long-distance scenarios. We filter a subset of episodes where the initial shortest-path distance exceeds 10 meters, comparing our approach against ApexNav on ObjectNav and UniGoal on TextInstanceNav. As illustrated in Fig.~\ref{fig:short-b}, our advantages are even more pronounced on these long-distance episodes compared to the full evaluation set, yielding average improvements of $9.4\%$ in SR and $5.0\%$ in SPL. These results demonstrate that our performance gains primarily stem from more effective handling of long-horizon navigation. By leveraging explicit hierarchical memory and evaluating long-term expected returns, the agent significantly mitigates meaningless exploration and redundant backtracking, thereby substantially improving path efficiency. More detailed results are provided in the supplementary material.

\begin{figure}[t]
\centering
\includegraphics[width=0.9\linewidth]{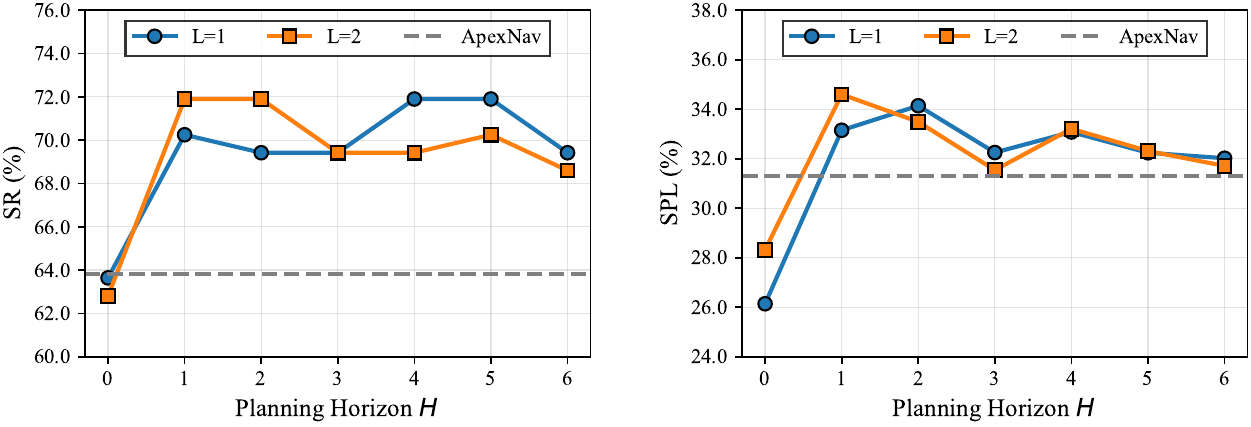}
\caption{
Level and planning horizon ablation. We evaluate the performance on long-range episodes of the HM3Dv2 ObjectNav dataset.
}
\label{fig:ablation_curve}
\end{figure}

\subsection{Ablation Study}

\textbf{Component Ablation.} In Table~\ref{tab:objnav_hm3dv2_ablation_components}, we remove exploration evidence, the zone-level hierarchy, long-horizon rollouts, and replace the VLM with Qwen3-VL-4B to isolate the impact of each component.

(i) w/o Evidence. Without exploration evidence, belief estimation relies solely on the LLM prior. Lacking online observation feedback to correct biases, the agent tends to over-search high-prior areas when the prior misaligns with the current scene.
(ii) w/o Zone-level. Simplifying the hierarchy to a coarse Object-Region structure restricts the planner's ability to leverage fine-grained spatial differences to filter the candidate space. 
(iii) w/o Rollouts. Without explicitly evaluating long-term returns  ($H=0$), the agent's greedy decisions lead to myopic exploration.
(iv) w/ Qwen3-VL-4B. Using a smaller VLM reduces the reliability of semantic descriptions and priors, which weakens the discriminative power of the belief state and subsequently degrades planning quality.

\noindent \textbf{Level–Horizon Ablation.} To investigate the interplay between hierarchy level and planning horizon, we evaluate combinations of level $L \in \{1,2\}$ and horizon $H \in \{0,1,2,3,4,5\}$ on long-range episodes of the HM3Dv2 ObjectNav dataset. $L=1$ and $L=2$ denote the Object--Region and Object--Zone--Region hierarchies, respectively. $H$ denotes the planning horizon,  i.e., the number of forward-simulation steps executed by the HSG simulator during online planning (Eq.~\ref{eq.q_value}). Results are shown in Fig.~\ref{fig:ablation_curve}.

\begin{figure*}[t]
    \centering
    \includegraphics[width=1.0\linewidth]{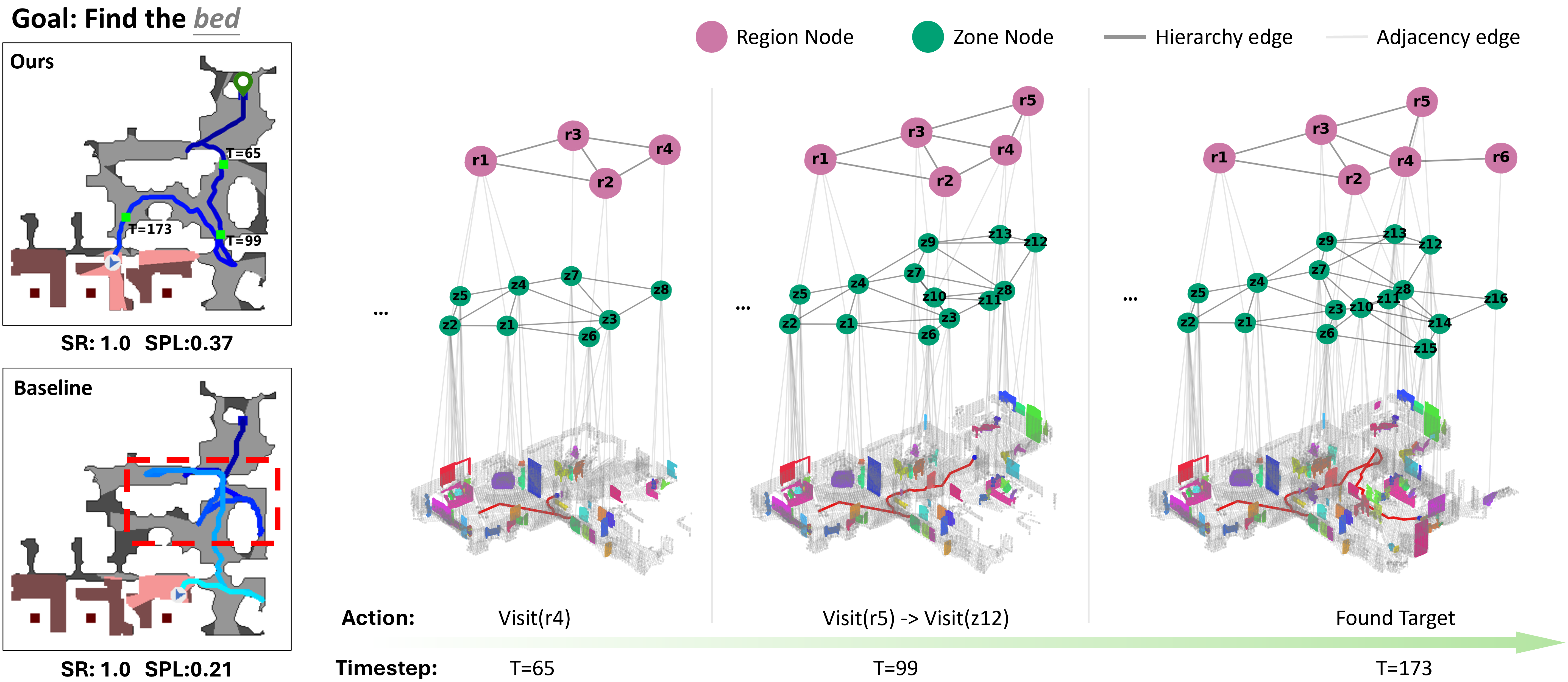}
    \caption{
    Visualization of the navigation process in a long-range episode. Left: top-down trajectories of our method and the baseline (ApexNav \cite{apexnav}) with SR/SPL; right: evolution of the HSG at three key timesteps, with the selected visit macro-actions and timesteps annotated.
    }
    \label{fig:nav_process}
\end{figure*}

\textit{The planning horizon $H$ admits an optimal value, and large $H$ can degrade performance.}  Increasing $H$ allows the planner to explicitly evaluate the long-term benefit of information gathering, improving long-range decision making. However, as $H$ grows, the mismatch between the HSG-based simulator and the real environment accumulates over multi-step simulation, making action-value estimates less reliable and leading to performance degradation.

\textit{The optimal horizon depends on the hierarchy level.} Since $L=2$ induces a more detailed HSG-based simulator, the multi-step simulation gap accumulates faster as $H$ increases, making longer planning horizons less reliable. As a result, $L=2$ prefers a shorter horizon than $L=1$.

\subsection{Qualitative Analysis}
Fig.~\ref{fig:nav_process} shows a representative long-range ObjectNav episode that illustrates our decision-making behavior. The baseline (ApexNav \cite{apexnav}) exhibits clear revisits in the mid-exploration stage (highlighted by the dashed red box), indicating redundant backtracking. In contrast, our policy plans at the zone/region level and evaluates macro-actions with long-term returns, avoiding redundant revisits.

\section{Conclusion}
In this paper, we propose an online hierarchical 3D scene graph construction and belief-based planning framework to address the core challenges of inadequate long-horizon decision-making and spatial memory in zero-shot semantic navigation. Operating in unseen environments, we integrate GVD with spectral clustering to incrementally construct a bottom-up, multi-granularity Object-Zone-Region semantic topology. This provides a structured state representation for global planning. Furthermore,  we fuse LLM semantic priors with exploration evidence to maintain hierarchical belief states. Through finite-horizon lookahead rollouts within the HSG-based simulator, the agent explicitly evaluates long-term expected returns of macro-actions. Extensive experiments show our approach outperforms SOTA baselines, especially in long-distance navigation tasks.


\noindent \textbf{Limitations and Future Work.} Our evaluation is limited to simulation and assumes posed RGB-D observations. In real-world deployment, localization errors may affect object-instance fusion and occupancy mapping. Moreover, although the system operates online, VLM/LLM inference introduces substantial latency and prevents real-time execution. Future work will focus on real-robot evaluation, integration with semantic SLAM, and more efficient semantic inference.

\section*{Acknowledgements}
This research is supported by Hong Kong Research Grants Council under GRF-15229423 and GRF-15221625.

%
%
\bibliographystyle{splncs04}
\bibliography{main}
\end{document}